# Modeling and Simulation of the Dynamics of the Quick Return Mechanism: A Bond Graph Approach

Anand Vaz[1*] and Thommen G.K.[2]

[1,2]Department of Mechanical Engineering, Dr. B.R. Ambedkar National Institute of Technology, Jalandhar

[*]Anand Vaz (email: anandvaz@nitj.ac.in)

## Abstract

This paper applies the multibond graph approach for rigid multibody systems to model the dynamics of general spatial mechanisms. The commonly used quick return mechanism which comprises of revolute as well as prismatic joints has been chosen as a representative example to demonstrate the application of this technique and its resulting advantages. In this work, the links of the quick return mechanism are modeled as rigid bodies. The rigid links are then coupled at the joints based on the nature of constraint. This alternative method of formulation of system dynamics, using Bond Graphs, offers a rich set of features that include pictorial representation of the dynamics of translation and rotation for each link of the mechanism in the inertial frame, representation and handling of constraints at the joints, depiction of causality, obtaining dynamic reaction forces and moments at various locations in the mechanism and so on. Yet another advantage of this approach is that the coding for simulation can be carried out directly from the Bond Graph in an algorithmic manner, without deriving system equations. In this work, the program code for simulation is written in MATLAB. The vector and tensor operations are conveniently represented in MATLAB, resulting in a compact and optimized code. The simulation results are plotted and discussed in detail.

## 1 Introduction

Bond graphs [1] are graphical tools which can be used to model and analyze the dynamic behavior of various multi-energy systems. The application of this technique results in a number of advantages, as described in [1, 2]. In addition, causality establishes a *cause* and *effect* relationship between the flow and effort elements of the bond. The notion of causality, apart from aiding with the formulation of system equations which govern the behavior of the system, help in pointing out any physical impossibility or system property we may have failed to take into account in the modeling stage.

Simulation has become an indispensible analytical tool using which one can experiment with a system at little or no expense. It is also used in design and development of appropriate control systems. In this work, a multibond or vector bond graph approach is used to model the dynamics of the quick return mechanism [3]. Using this Bond Graph model, we simulate the dynamic behavior of the system with appropriate codes written in MATLAB [4]. This establishes a very effective method of predicting the behavior of systems, which, in a number of applications can prove to be economical as well as time saving.

In general, mechanical systems can be treated as a finite number of rigid bodies, interconnected suitably, with appropriate constraints imposed at the joints. Kinematics of such mechanisms is usually available in most texts and reference books on machines and mechanisms [1, 5]. However, the dynamics is rarely presented although it is extremely important from the design perspective. Using the Bond Graph approach, simulation can be conveniently carried out in MATLAB, and the dynamic quantities like reaction forces and torques at various locations on the system can be determined, plotted and analyzed. Also, the actuator forces and torques required to produce desired behavior can be determined.

The primary aim of this work is to demonstrate the modeling of the dynamics of the quick-return mechanism and then discuss the results obtained using a realistic computer model of this system to simulate its response by means of appropriate programs written in MATLAB. The system equations derived from the bond graphs are in a form which can directly be used to write the program code for the simulation of the system in MATLAB. The results of the simulation are then discussed in detail.

## 2 Modeling

In this work, we develop a multibond graph model of the quick return mechanism, representing the translation and rotation for each rigid link of the system. The quick return mechanism is treated as one comprising of five rigid moving links, having relative motion with respect to each other, and also with respect to a stationary **0** frame. This inertial frame does not possess any translational or rotational motion. The center of mass of each link is assumed to be located at its corresponding geometric center. The translational effect is concentrated at the center of mass of each link, while the rotational effect is considered in the inertial frame itself by considering the inertia tensor for each link about its respective center of mass and expressed in the inertial frame. Reference frames are fixed on each link, using the Denavit-Hartenberg convention [6]. Fig. (1) shows the different links of the system, along with their corresponding centers of mass and associated frames. The links are interconnected with each other by imposing suitable constraints at the joints. A problem of *differential causality* [1, 2] arises while modeling the joints, which is rectified by introducing suitable stiffness and damping elements while imposing constraints. These elements make the model more realistic by bringing in the effects of compliance and dissipation at the joints, within definable tolerance limits. The multibond graph is then causaled and the codes for simulation in MATLAB are directly derived from it. Bond graph modeling of rigid multibody systems has been presented using scalar bond representations [7, 8], and using multibond graphs [9-15]. However, a clear approach derived from first principles has been presented in [14]. The simulation results obtained are then plotted, analyzed and discussed in detail.

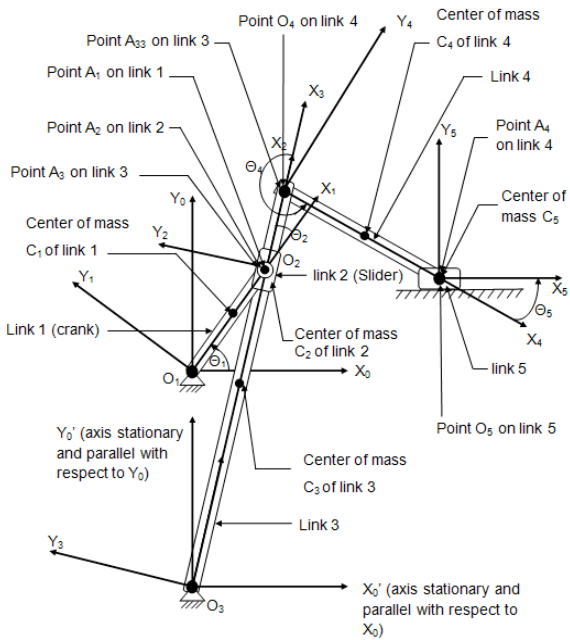

Fig. 1: The Quick Return Mechanism

The mechanism has five moving links with reference frames fixed to each link. The centers of mass of each of the links are also shown. $X_0$-$Y_0$ and $X_0'$-$Y_0'$ are inertial frames of reference.

## 2.1 The Quick Return Mechanism

The quick return mechanism is used to convert the rotational motion of the crank to translational motion of the slider. The return stroke of the slider takes a lesser span of time to complete, as compared to the forward stroke. The individual components of the mechanism are considered as rigid bodies [13-14], connected at the joints. The mechanism consists of a crank (link **1**), connected to a slider (link **2**) which slides along a rocker arm (link **3**). The rocker arm is connected to the final sliding link (link **5**) through a connecting rod (link **4**). Reference frames are fixed on each link, that is, frame **1** is fixed on link **1**, frame **2** on link **2** and so on. A fixed inertial frame **0**, whose origin coincides with that of frame **1**, is chosen. However, it will neither translate nor rotate. $C_1$, $C_2$, $C_3$, $C_4$ and $C_5$ are the centers of mass of the respective links. As mentioned earlier, the Denavit-Hartenberg convention is used to fix the reference frames on each of the links. The dynamics of the system of fig. (1) is modeled in the multibond graph shown in fig. (2). The model depicts rotational as well as translational dynamics for each link in the system. The left side of the bond graph is associated with the rotational dynamics of the system, while the right side represents the translational dynamics. We restrict any relative translational motion between the origin of the inertial frame **0** and point $O_1$ on link **1** by applying source of flow $S_f$ as zero. Similarly, the relative motion at point **A**, distinguished by $A_1$ on link **1**, $A_2$ on link **2** and $A_3$ on link **3**, is restricted by applying a source of flow $S_f$ equal to zero. A similar approach is used to model the joints between link **3** and **4,** at point $A_{33,}$ and between link **3** and the inertial frame, at point $O_3$. The slider, which is link **2**, is constrained to translate along the $X_3$ direction only. In order to constrain the motion of link **2** in the moving frame **3**, all the flows and efforts which were initially expressed in the inertial frame are expressed in frame **3**. This is achieved by using a modulated transformer, having a modulus equal to $-\begin{bmatrix}^{0}_{3}R\end{bmatrix}$.

*Differential causality* which arises at the joint of the sliding link (link **2**) is eliminated by setting the $K_{3C}(1,1)$ element in the stiffness matrix $[K_{3C}]$ between frames **0** and **3** as zero. The multibond is split into its scalar components and the translation along $Y_3$ and $Z_3$ directions is constrained by applying source of flow $S_f$ equal to zero for these components. A similar treatment is carried out while modeling the joint at the second slider (link **5**). It is also observed that link **2** and link **3** have the same angular velocity, as seen from the inertial frame. Thus, in order to constrain any relative angular motion between link **2** and link **3,** we apply a source of flow $S_f$ equal to zero, between the rotational sides of these two links. The viscoelastic elements used at the joints are represented by **C** and **R** elements. For the simulation, the crank is made to rotate about point $O_1$ with an angular velocity equal to 5 rad/s. This is achieved by applying a source of flow $S_f$ = 5 rad/s to the fixed end of link **1** about the **z** direction.

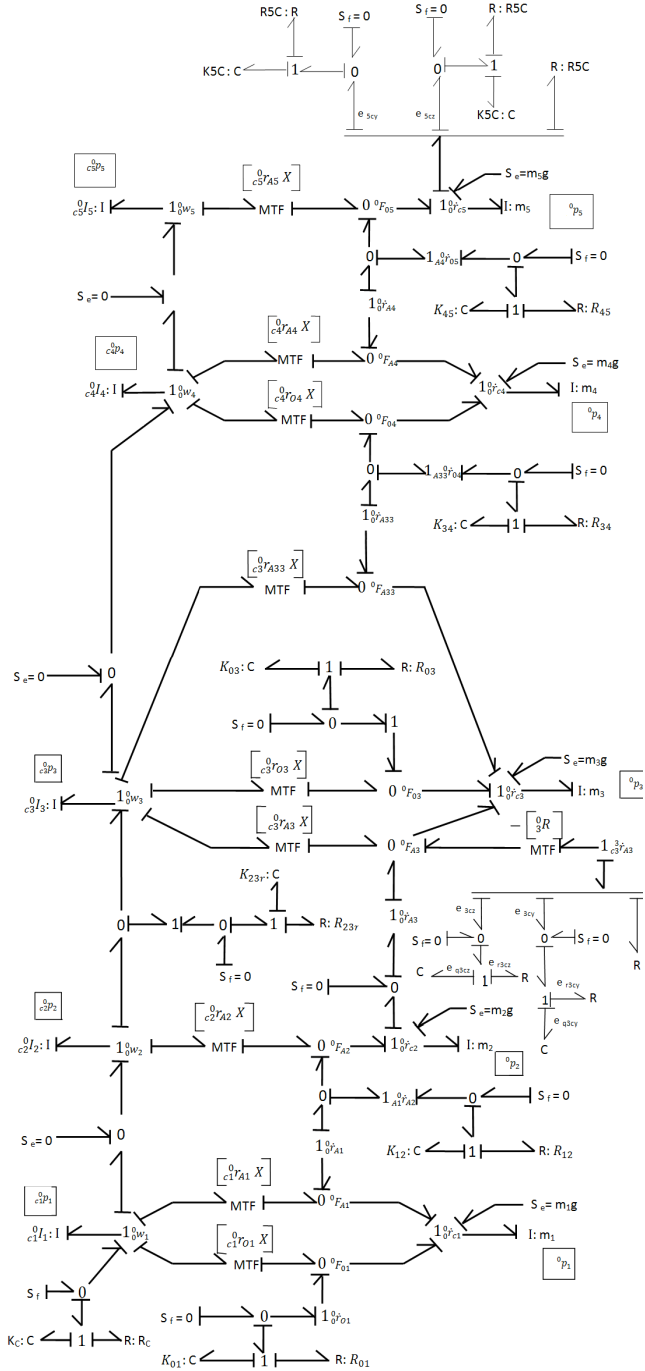

Fig. 2: Multibond graph model of the quick return mechanism

In this model, the inertia elements are denoted by I, and the stiffness and damping elements by K and R respectively.

## 3   Simulation Results

The Bond Graph model has been simulated using MATLAB. The simulation results are shown in this section. Table 1 shows the values of the various link parameters used in the simulation, whereas table 2 shows the values of the stiffness and damping of the various couplings used.

**Table-1:** Link parameters used for simulation

|  | Crank | Slider | Rocker Arm | Connecting Rod | Slider 2 |
|---|---|---|---|---|---|
| Mass | 0.5 kg | 0.1 kg | 0.7 kg | 0.3 kg | 0.1 kg |
| Length along x-axis | 0.2 m | 0.01 m | 0.7 m | 0.4 m | 0.01 m |
| Length along y-axis | 0.01 m | 0.01 m | 0.01 m | 0.01 m | 0.01 m |
| Length along z-axis | 0.01 m | 0.01 m | 0.01 m | 0.01 m | 0.01 m |

**Table-2:** Values of stiffness and damping of different couplings used for simulation

|  | Stiffness, K | Damping, R |
|---|---|---|
| Translational coupling between crank and fixed frame | $K_{01}$ = 100000 N/m | $R_{01}$ = 20 N·s/m |
| Translational coupling between slider and crank | $K_{12}$ = 100000 N/m | $R_{12}$ = 20 N·s/m |
| Translational coupling between rocker arm and fixed frame | $K_{03}$ = 100000 N/m | $R_{03}$ = 20 N·s/m |
| Translational coupling between slider and rocker arm | $K_{3C}$ = 100000 N/m | $R_{3C}$ = 20 N·s/m |
| Rotational coupling between slider and rocker arm | $K_{23r}$ = 100 N·m/rad | $R_{23r}$ = 0.5 N·m·s/rad |
| Rotational coupling between source of constant angular velocity and crank | $K_C$ = 100 N·m/rad | $R_C$ = 100 N·m·s/rad |

| Translational coupling between rocker arm and connecting rod | $K_{34}$ = 100000 N/m | $R_{34}$ = 20 N·s/m |
|---|---|---|
| Translational coupling between connecting rod and slider 2 | $K_{45}$ = 100000 N/m | $R_{45}$ = 20 N·s/m |
| Translational coupling between slider 2 and fixed frame | $K_{5C}$ = 100000 N/m | $R_{5C}$ = 20 N·s/m |

### 3.1 Simulation Plots

The simulation plots for the different links have been discussed in the following sections.

#### 3.1.1 Dynamics of the Crank

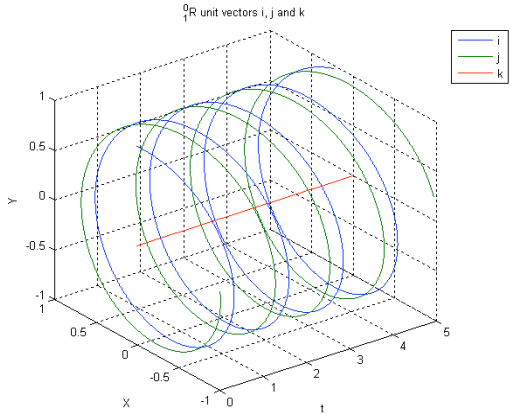

Fig. 3: Variation of orientation matrix of link 1 with time

The crank rotates in the **x-y** plane. The orientation matrix $\begin{bmatrix}^0_1R\end{bmatrix}$ represents the projection of unit vectors of frame **1** (the crank) on frame **0** (the stationary frame). The unit vectors in **x** and **y** directions move in a circular path, while the unit vector in the **z** direction is stationary. Fig. (3) shows the variation of the unit vectors with time in a 3D plot.

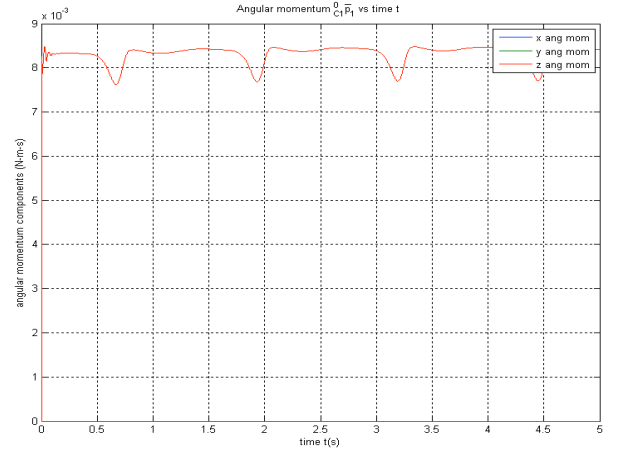

Fig. 4: Variation of angular momentum of link 1 with time.

As indicated in the Fig. (4), the **z** component of angular momentum is almost constant with time. This is due to the constant value of source of flow $S_f$ imposed on the crank. The small variation can be explained as effects of the compliance element $K_C$. The initial transients which arise due to the sudden imposition of source of flow $S_f$ die down after a brief initial period due to damping.

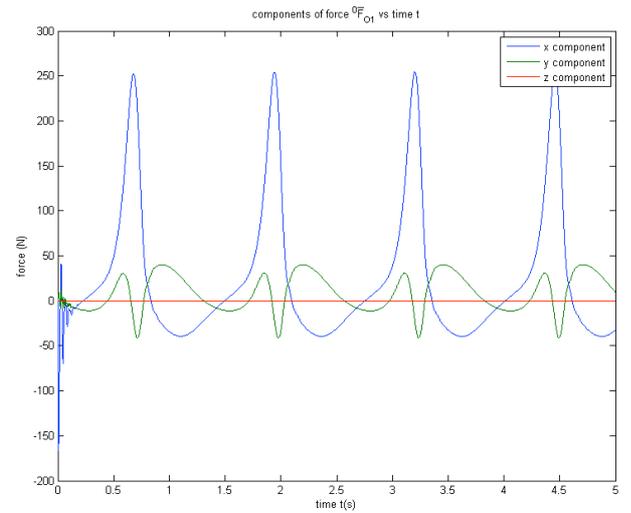

Fig. 5: Variation of force at fixed end of crank with time

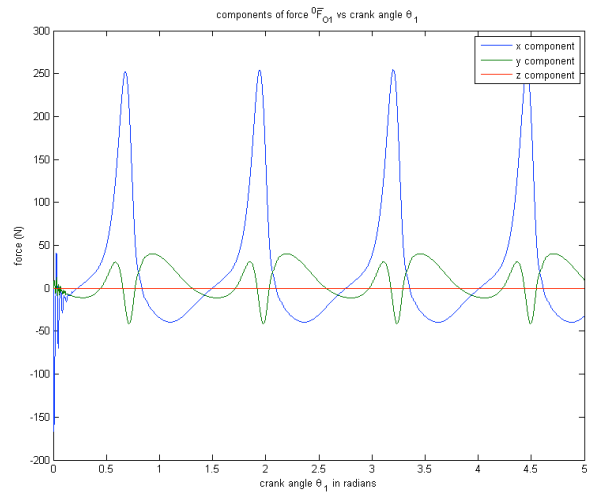

Fig. 6: Variation of force at fixed end of crank with crank angle

The crank is pivoted at the point **$O_1$**. As the crank rotates, the motion of the crank causes forces to be transmitted on the fixed end. Fig. (5) shows the variation of these forces with time and fig. (6) shows the variation with crank angle. In both the figures, we observe initial transients for a short period after the system is set into motion. These transients occur due to sudden application of source of flow $S_f$ = [0 0 5]' rad/s on the crank from t = 0. In the simulation, an initial crank angle of 90° or 1.57 radians is given. So, in fig. (6), the transients start at an angle of 1.57 radians, that is, the angle from which the crank starts rotating. The forces reach a maximum value during this initial period and gradually, the transients die down and thereafter, force variation continues in a periodic manner. The **x**, **y** and **z** components of the force vector are plotted separately. The plot lines at the ends show a discontinuity. This is perhaps not part of the response, but due to the angle exceeding one revolution. These force plots can be used as a guideline for design purposes, as these represent the actual dynamic reaction forces at the crank shaft axis during its working.

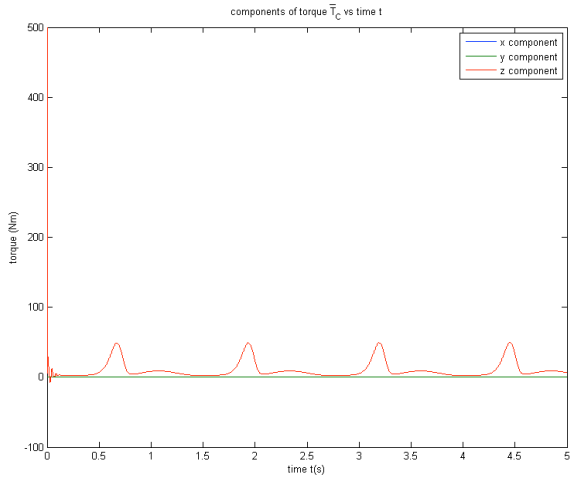

Fig. 7: Variation of torque Tc on the crank with time

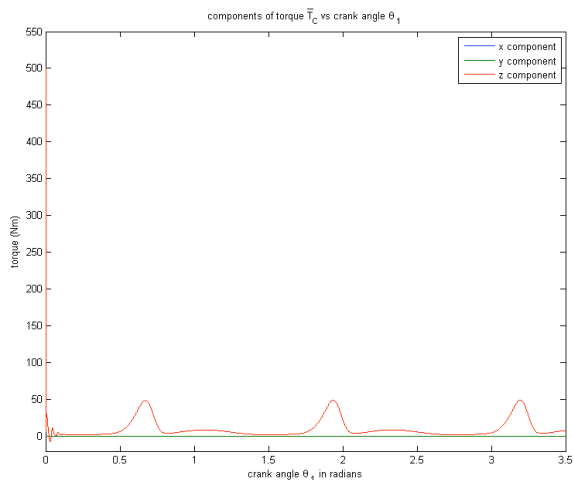

Fig. 8: Variation of torque Tc on the crank with crank angle

We have considered a constant source of flow of 5 rad/sec at the crank. This source of flow generates a torque which rotates the crank and hence transmits forces to the rest of the links to actuate the entire mechanism. Fig. (7) and fig. (8) indicate the variation of this generated torque with time and crank angle respectively. The crank starts from an initial angle of 90° or 1.57 radians. As seen in the force plots, we observe initial transients in the torque plots. This is observed at 1.57 radians in fig. (8), due to the sudden imposition of a source of flow. Torque is maximum at 1.57 radians. This is required to start the motion of the system from its initial position of rest. These transients gradually die down and we observe the forced part of the response. Thus, the actuator specifications can be determined.

### 3.1.2 Dynamics of the Slider

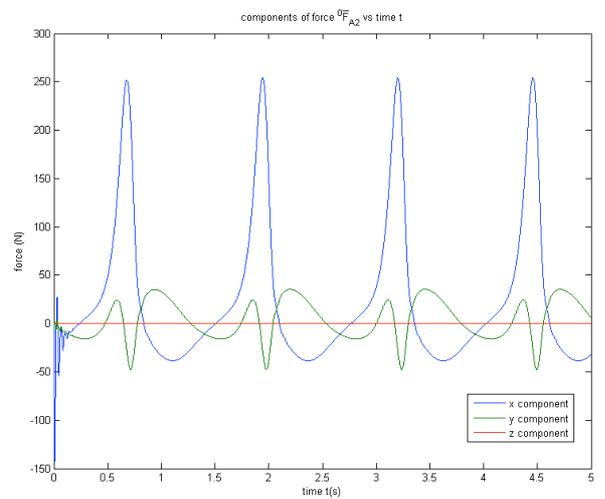

Fig. 9: Variation of force at pin connecting crank and slider at point **$A_2$** on slider with time

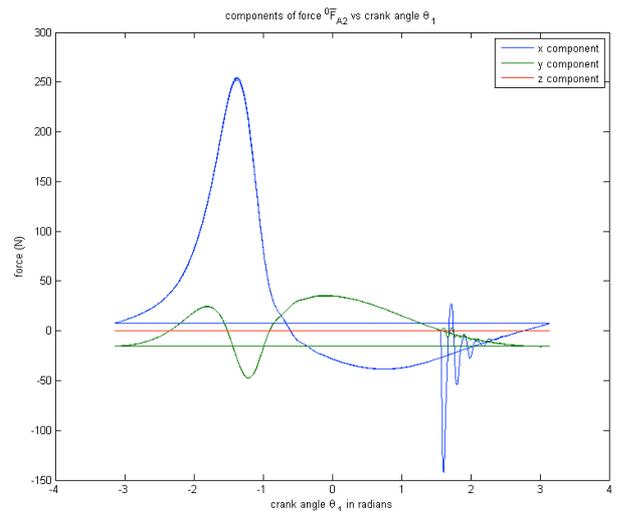

Fig. 10: Variation of force at pin connecting crank and slider at point **$A_2$** on slider with crank angle

The point at which the crank is connected to the slider is point **$A_2$**. Due to interaction with other links such as the slider, forces get transmitted to the point $A_2$. Fig. (9) indicates the variation of the forces with time and fig. (10) indicates the variation of forces with crank angle.

Transients exist for a short period after the system is set into motion, after which it gradually dies down due to damping. We observe the forced response of the system thereafter. The plot lines at the ends show a discontinuity due to the angle exceeding one revolution. The **z**-component is zero and the forces exist in the **x-y** plane only.

### 3.1.3 Dynamics of the Rocker Arm

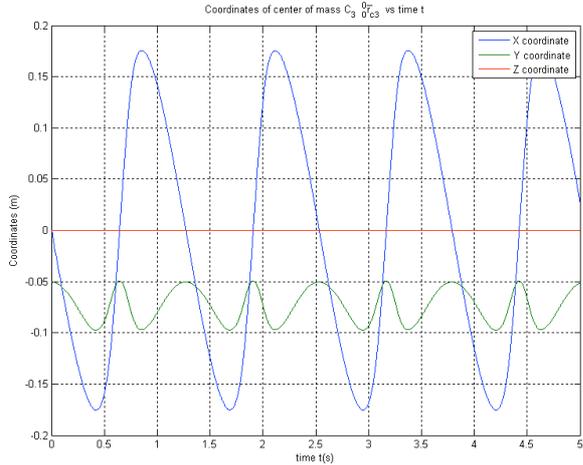

Fig. 11: Variation of coordinates of center of mass of rocker arm with time

The center of mass of the rocker arm oscillates about the **z**-axis. Fig. (11) indicates the variation of the components of the position vector $^0_0\bar{r}_{C3}$ of center of mass of the rocker arm. The **z**-component is constant as the center of mass of the rocker arm oscillates in the **x-y** plane, about the **z**-axis. From fig. (11), it can be observed that the time variation of the position of the **x** and **y** coordinates of the center of mass is steeper during the return stroke and less steep during the forward stroke. This implies that the return stroke takes a shorter duration to complete as compared with the forward stroke. This is due to the quick-return action of the mechanism.

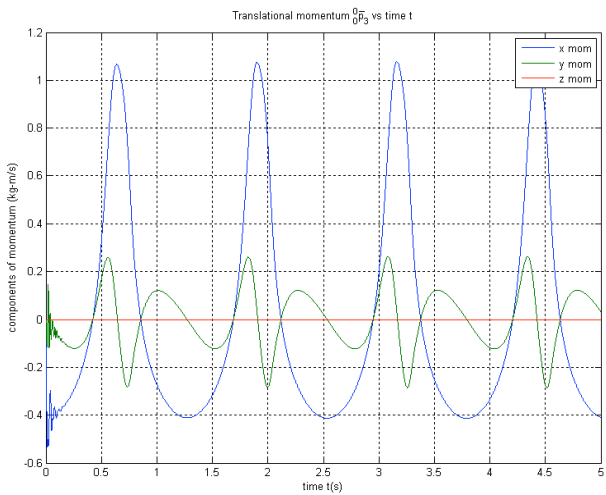

Fig. 12: Variation of translational momentum of rocker arm with time

The variation of the translational momentum of the rocker arm is indicated in fig. (12). The **x**-component and **y**-component of the translational momentum vector vary with time and the **z**-component is 0, as the rocker arm rotates about the **z**-axis. Transients are observed at the beginning of the plot. These transients occur due to sudden imposition of the source of flow, or motion, on the crank. As observed in the previous plots, the transients die down after some time and we observe a periodic variation of the translational momentum, as shown in the figure, which represents the forced response of the system. An interesting observation is that the peak values of the translational momentum for both **x** and **y** coordinates, have greater magnitude during the return stroke (part of the graph with positive values of momentum) as compared with the forward stroke (part of the graph with negative values of momentum). This shows the quick return phenomenon.

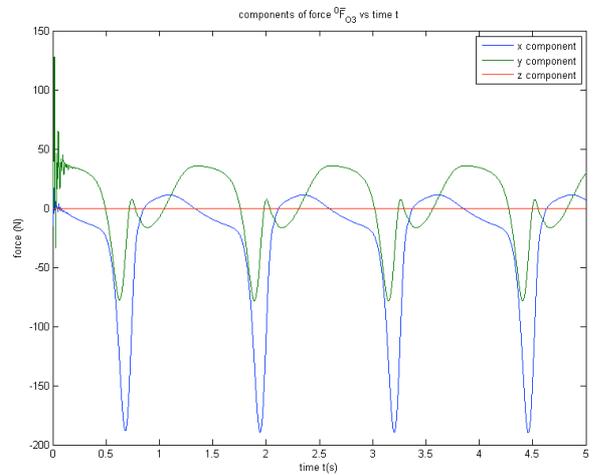

Fig. 13: Variation of force at fixed end of rocker arm with time

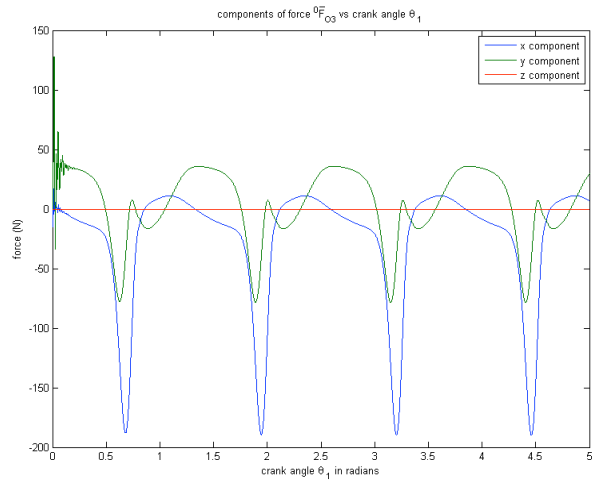

Fig. 14: Variation of force at fixed end of rocker arm with crank angle

The rocker arm is pivoted at the point $O_3$. As link **3** is set into motion by the crank, forces develop at the point $O_3$. The force at $O_3$ is plotted in fig. (13) and fig. (14). Fig. (13) and fig. (14) shows the variation of the force with time and with crank angle respectively. As observed in the other force plots, the initial transients gen-erated due to a sudden imposition of a source of flow, dies down after a period of time. The maximum forces that occur at the point $O_3$ can be determined and the system can be designed and materials be decided accordingly.

### 3.1.4 Dynamics of the Connecting Rod

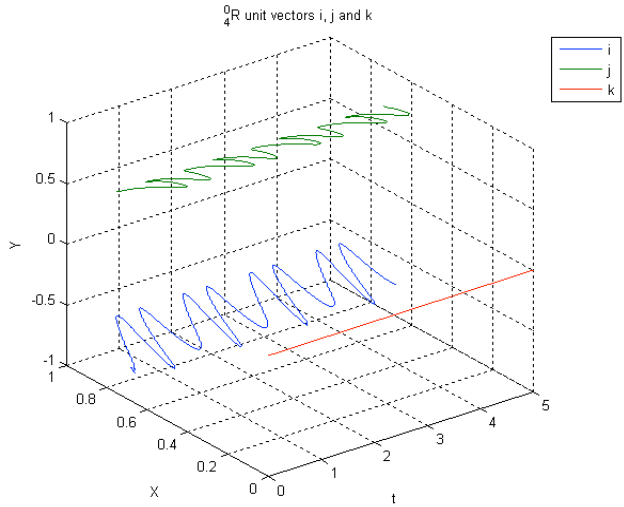

Fig. 15: Variation of orientation matrix of link 4 with time

The orientation matrix $\begin{bmatrix} ^0_4R \end{bmatrix}$ represents the projection of unit vectors of frame **4** (the connecting rod) on frame **0** (the stationary frame). The unit vectors in **x** and **y** directions move in a circular arc, while the unit vector in the **z** direction is stationary. Fig. (15) shows the variation of the unit vectors with time in a 3D plot.

### 3.1.5 Dynamics of Slider 2 (Link 5)

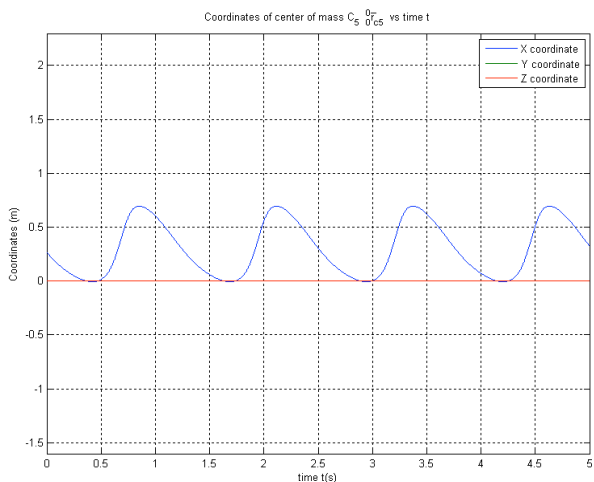

Fig. 16: Variation of coordinates of center of mass of slider 2 with time

The center of mass of the slider 2 reciprocates along the **x** direction. The **y**-component and **z**-component is constant as the center of mass of the slider 2 reciprocates along the **x**-axis. From fig. (16), it can be observed that the time variation of the position of the **x** co-ordinates of the center of mass is steeper during the return stroke and less steep during the forward stroke. This is due to the quick-return action of the mechanism.

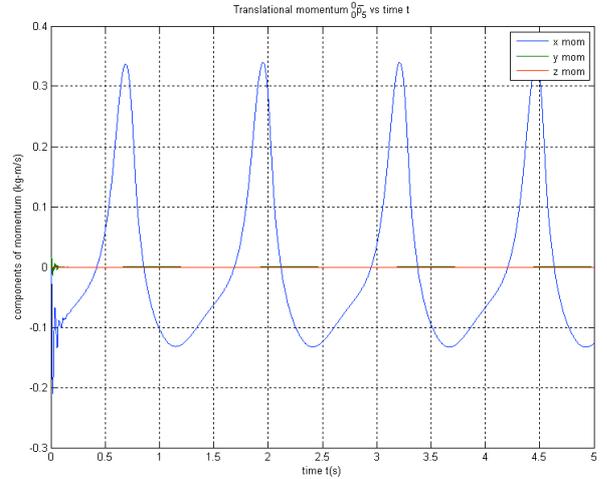

Fig. 17: Variation of the components of translational momentum with time

As seen from the previous graphs, it is seen that the slider 2 reciprocates in the **x** direction. Its motion in the **y** and **z** directions is constrained. This is reflected in fig. (17) which shows the variation of the **x** component of momentum with time. It is seen that the slider 2 attains higher peak values of the **x** component of translational momentum during the return stroke as compared to the peak values in the forward stroke. This is due to the fact that it has a greater velocity during the return stroke as compared to the forward stroke, due to the quick return action of the mechanism.

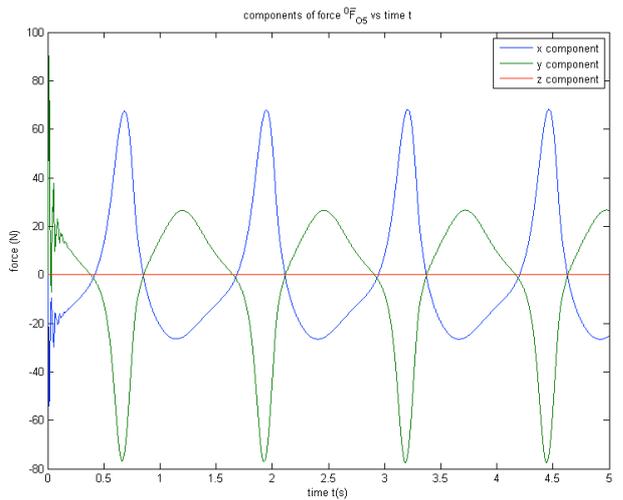

Fig. 18: Variation of force at the center of mass of slider 2 with time

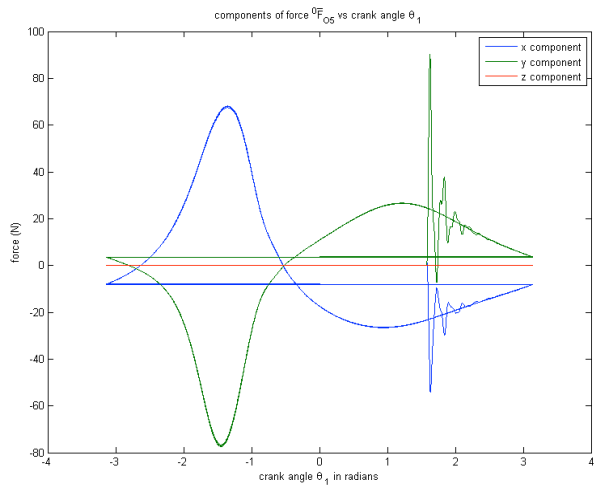

Fig. 19: Variation of force at center of mass of slider 2 with crank angle

This point $O_5$ is a point on the slider 2 that moves along parallel to the $X_5$ axis. As the crank rotates, due to the interaction of link **5** with the other links, forces are transmitted to point $O_5$. Fig. (18) shows the variation of these forces with time and fig. (19) shows the variation of the forces with crank angle. As seen in the force plots of the other links, initial transients can be observed. The initial transients die down and thereafter, the force variation is found to be repetitive. The maximum force occuring at this point can be determined and the joint can be accordingly designed.

## 4   Conclusions

The quick return mechanism has been modeled and its dynamics simulated in MATLAB using the Bond Graph approach. The codes for simulation have been directly obtained from the Bond Graph model. This approach has not only enabled us to conveniently extract information regarding the kinematic aspects of each link of the mechanism, but also the dynamic aspects, which is very important from the design point of view.